\documentclass[letterpaper, 10pt, conference]{ieeeconf}      

\IEEEoverridecommandlockouts                              
\renewcommand{\baselinestretch}{1.0}

\usepackage{nicefrac}
\usepackage{url}
\usepackage{subfig}
\usepackage[dvipsnames]{xcolor}
\usepackage{multirow}
\usepackage{bbold}
\usepackage{balance}

\usepackage{breakurl}

\usepackage{footnote}
\makesavenoteenv{tabular}
\makesavenoteenv{table}

\def\HiLi{\leavevmode\rlap{\hbox to \hsize{\color{yellow!20}\leaders\hrule height .8\baselineskip depth .5ex\hfill}}}
\def\HiLiGreen{\leavevmode\rlap{\hbox to \hsize{\color{green!20}\leaders\hrule height .8\baselineskip depth .5ex\hfill}}}
\def\HiLiBlue{\leavevmode\rlap{\hbox to \hsize{\color{blue!20}\leaders\hrule height .8\baselineskip depth .5ex\hfill}}}
\def\HiLiGray{\leavevmode\rlap{\hbox to \hsize{\color{grey!50}\leaders\hrule height .8\baselineskip depth .5ex\hfill}}}

\usepackage{epsfig}
\usepackage{amssymb}
\usepackage{amsmath}
\usepackage{amsfonts}

\usepackage{stackengine}
\usepackage{algorithmic,comment}
\usepackage[ruled,vlined,commentsnumbered,titlenotnumbered,linesnumbered]{algorithm2e}
\usepackage{float}

\usepackage{colortbl}
\usepackage{xcolor}
\definecolor{LightCyan}{rgb}{0.88,1,1}

\usepackage{color}
\newcommand{\SC}[1]{{\color{black}#1}}

\newcommand{\tu}[1]{{\underline{\textit{#1}}}}

\title{\LARGE \bf
Personalized Federated Learning of Driver Prediction Models for Autonomous Driving}

\author{Manabu Nakanoya$^{1}$, Junha Im$^{2}$, Hang Qiu$^{3}$, Sachin Katti$^{3}$, Marco Pavone$^{3}$, Sandeep Chinchali$^{4}$
\thanks{$^{1}$ NEC Corporation, Kanagawa, Japan {\small \tt{nakanoya@nec.com}}}
\thanks{$^{2}$ Samsung Electronics, Hwaseong, Korea {\small \tt{junha130.im@samsung.com}}}
\thanks{$^{3}$ Stanford University, Stanford, CA, USA \tt{\{hangqiu, skatti, pavone\}@stanford.edu}}
\thanks{$^{4}$ The University of Texas at Austin, Austin, TX {\small \tt{sandeepc@utexas.edu}}}
}

\begin{document}
\maketitle

\begin{abstract}
Autonomous vehicles (AVs) must interact with a diverse set of human drivers in heterogeneous
geographic areas. Ideally, fleets of AVs should share trajectory data to continually re-train and improve
trajectory forecasting models from collective experience using cloud-based distributed learning. At the same time,
these robots should ideally avoid uploading raw driver interaction data in order to protect proprietary policies (when sharing insights with other companies) or protect driver privacy from insurance companies. 
Federated learning (FL) is a popular mechanism to learn models in cloud servers from diverse users without divulging private local data. However, FL is often not robust -- it learns sub-optimal models when user data comes from highly heterogeneous distributions, which is a key hallmark of human-robot interactions. In this paper, we present a novel variant of \textit{personalized} FL to specialize robust robot learning models to diverse user distributions. Our algorithm outperforms standard FL benchmarks by up to $\SC{2 \times}$ in real user studies that we conducted where human-operated vehicles must gracefully merge lanes with simulated AVs in the standard CARLA and CARLO AV simulators. 

\end{abstract}

\section{Introduction}
\label{sec:introduction}

Future robotic fleets must operate amongst diverse humans with heterogeneous
preferences on human-robot interaction, in applications ranging from nursing assistance robots to home robots and autonomous vehicles (AVs).
Given such heterogeneity, there is a large incentive to share data from robotic fleet deployments to improve computer vision, prediction, and control modules based on diverse interactions with human users. For example, AVs can learn human trajectory forecasting models to proactively anticipate the behavior of nearby humans to aid decision-making \cite{schmerling2018multimodal}. Sharing rare and risk-sensitive driver styles during challenging contexts (e.g. traffic disruptions) can potentially help learn more robust forecasting models that generalize to new cities and driver populations. At the same time, raw trajectory data should be kept private to avoid revealing proprietary logic to competitor companies or individual driver behavior to insurance companies. In this paper, we address how to balance the competing objectives of user privacy and utility of data sharing in large-scale robotic fleet learning. 

Federated learning (FL) is a promising approach to train machine learning (ML) models from distributed datasets while preserving privacy \cite{google2017federated,mcmahan2016federated,li2020federated}. FL trains a model locally at a user's device (e.g. a robot) and simply shares model parameter updates with a central server to learn from diverse users while protecting privacy by not uploading raw training data. Typically, FL performs poorly (e.g., converges to a poor global model) when user data
is highly heterogeneous \cite{kairouz2019advances,fallah2020personalized}, which is a key problem for AVs that must interact with a variety of diverse human driving styles. The recently-proposed \textit{personalized} variant of FL \cite{fallah2020personalized,dinh2020personalized} is a promising approach to handle user diversity, but, to the best of our knowledge, has hitherto not been applied in robotics. 
Moreover, our subsequent results illustrate poor performance for standard FL in robotics applications owing to diverse human-robot interactions. As such, 
the key contribution of this work is to contribute novel algorithms for personalized FL for human trajectory forecasting models in AV deployments.


\textit{Contributions and Organization: } The technical contributions and organization of this paper are as follows. To the best of our knowledge, we present the first user study that assesses the efficacy of \textit{personalized} federated learning in robotics, especially for trajectory forecasting models for AVs. Then, we introduce a simple experiment where a fleet of robots have similar dynamics models, but widely different cost functions, for which standard applications of personalized FL
are sub-optimal. Then, to mitigate these problems, we introduce novel algorithmic extensions to personalized FL that effectively learn from global experience for common parameters (e.g., shared dynamics models) while adaptively specializing local cost functions.
Finally, we show strong performance gains for our algorithm in a user-study with real human drivers on photo-realistic simulators like CARLA \cite{Dosovitskiy17} and lightweight simulators like CARLO \cite{cao2020reinforcement} for a lane merging scenario requiring challenging human-robot interaction.

\begin{figure}[t]
    \centering
    \subfloat{
    \includegraphics[width=0.48\columnwidth]{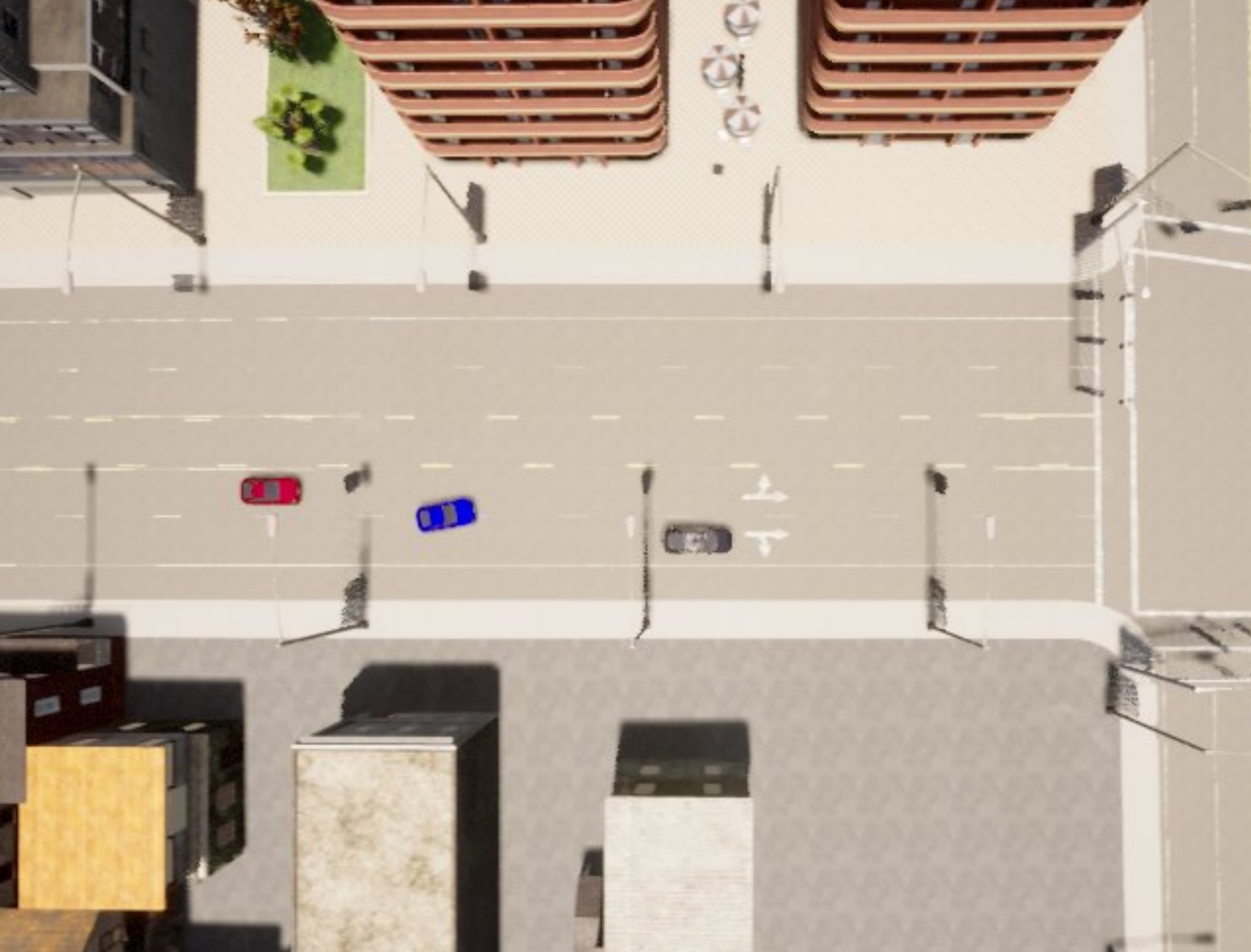}
    }
    \subfloat{
    \includegraphics[width=0.48\columnwidth]{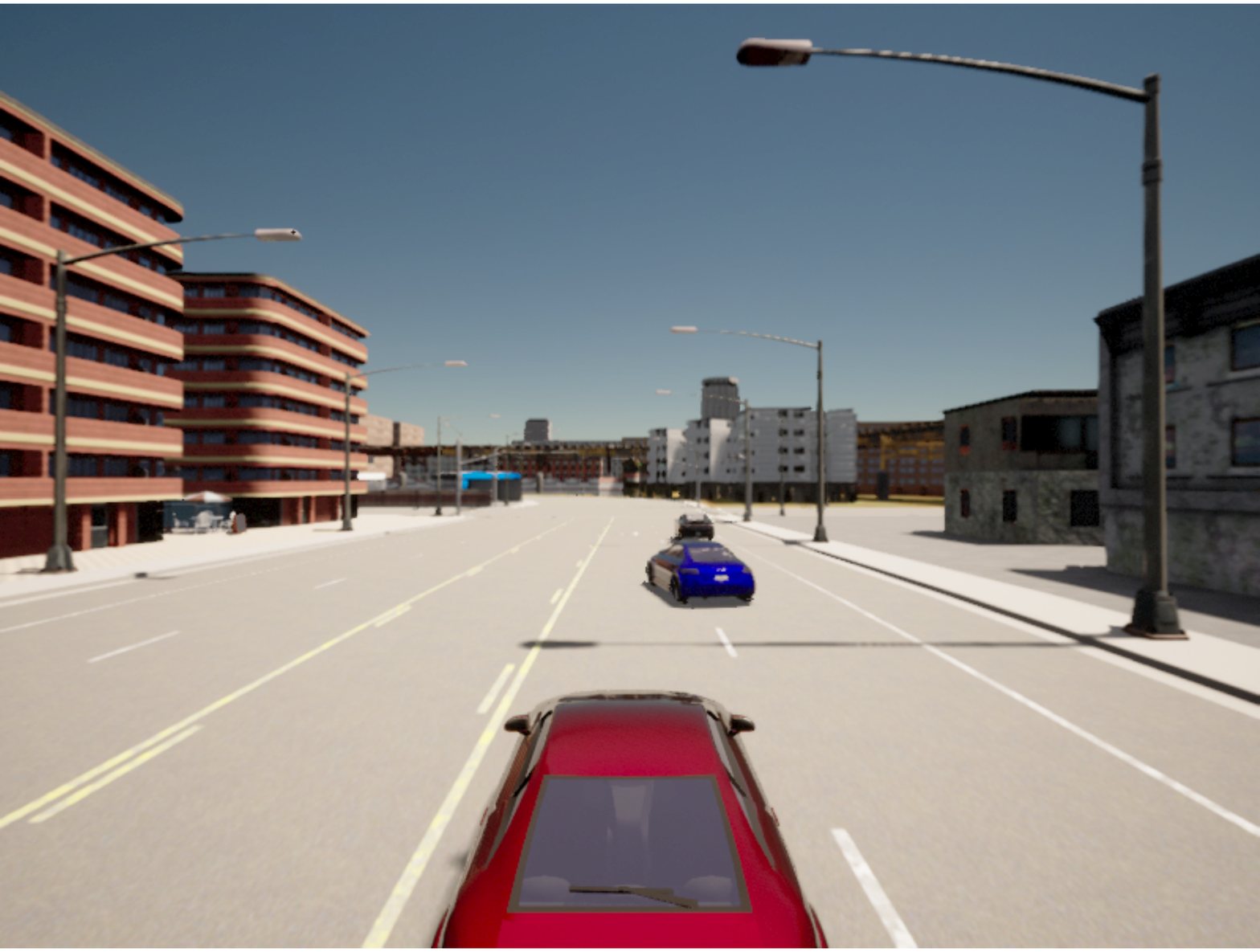}
    }
    \caption{\small{We evaluate a lane-merging scenario in CARLA with autonomous and human-driven vehicles to test the efficacy of \textit{personalized} federated learning of trajectory forecasting models.}}
    \label{fig:carla_screen_shot}
\vspace{-1em}
\end{figure}

\section{Related Work}
\label{sec:related_works}
\textit{Motivation for Personalized Federated Learning in Robotics: }
FL \cite{google2017federated} is a method to train a global ML model from multiple networked devices, such as mobile phones, that each have local labelled datasets. The key benefits of FL are privacy protection and reduced communication overhead since locally-trained model parameters, instead of raw user data, are sent to the cloud for knowledge sharing. FedAVG \cite{mcmahan2016federated} is a simple, widely-used implementation of the FL algorithm that averages model gradients from diverse clients to learn a global model, which is then synced back to individual clients for continual learning. 
Standard applications of FL, including FedAVG, lose robust performance when individual mobile clients' (e.g., robots) data is highly heterogenous \cite{kairouz2019advances,li2020federated,bagdasaryan2020backdoor}.
Such vulnerability to dataset heterogeneity is a crucial problem for networked robotic fleets such as AVs, 
since perception and forecasting models can fail to converge when robots observe sensory data from diverse environment distributions and interactions with radically different humans.  

Recently, personalized FL \cite{deng2020adaptive,dinh2020personalized,fallah2020personalized} has emerged as an effective method to learn robust models under such heterogeneity by specializing potentially unique models for each client (e.g. robot).
It aims to first learn a global model and then efficiently adapt it to individual robots while minimizing the extra training cost of personalization. This technique is inspired by multi-task learning \cite{ruder2017overview,zhang2017survey} and meta-learning \cite{vilalta2002perspective}, and is especially important in situations where the cost of personalization is relatively high, such as for low-power robots or mobile phones.
Another approach to personalized FL is to weigh the parameters of a global model and unique personalized local model. 
For example, Deng et al. \cite{deng2020adaptive} propose an algorithm that adaptively changes the relative weighting of parameters of a global model during local training. This approach is generally able to improve the accuracy of personalized models, although it leads to increased computation and communication costs.

Our key observation is that many models in robotics have common internal structures -- dynamics models are often shared by similar robots, but cost functions or risk sensitivities can be unique. As such, our approach embraces the fact that robotic models can have a subset of parameters that are relatively invariant to data measured on each robot (e.g., shared dynamics) and should therefore resemble global parameters. On the other hand, other subsets of parameters should be personalized for each client, such as those that model risk-sensitivity and cost functions. Thus, our method flexibly adjusts the learning rate for each class of parameters based on its variance across heterogenous robots, which leads to more accurate and stable learning.
As such, we embrace that different sub-sets of parameters represent global and local patterns, which makes our work different from \cite{deng2020adaptive} which simply produces a weighted average of global and local models.

\textit{Adaptive Learning Rates in Optimization: }
In optimization, the learning rate is a hyper-parameter that governs how quickly parameters can be updated during one step of gradient descent, such as 
in the standard Stochastic Gradient Descent (SGD) algorithm.
For example, the AdaGrad \cite{duchi2011adaptive} algorithm adjusts the learning rate for each parameter based on its cumulative gradient. This algorithm helps to stabilize and speed up the learning process because it can apply a more appropriate learning rate than a uniform one depending on the progress of learning individual parameters. SGD algorithms such as RMSProp \cite{tieleman2012lecture} and Adam \cite{kingma2014adam}, which are widely used in modern ML training, also use this core idea, which can easily be used in personalized FL. In key contrast to these algorithms, our algorithm adjusts the learning rate 
based on the progress of learning \textit{across distributed robots}. Specifically, our algorithm calculates the variance of each individual parameter across FL clients (robots) to gauge the progress of learning and set an appropriate learning rate for that parameter. Moreover, our method can be used alongside standard SGD optimizers for local datasets, such as Adam, as we show in our evaluation.

\textit{Trajectory Forecasting for AVs: }
Our work is complementary to a rich body of research on trajectory forecasting models in robotics, which attempt to predict the motion of pedestrians
or the future trajectories of human drivers conditioned on behaviors of AVs.
For example, Gupta et al. \cite{gupta2018social} proposes a Generative Adversarial Network (GAN) model that predicts pedestrians trajectories while considering their interactions. Likewise, Ivanovic et al. \cite{ivanovic2019trajectron,salzmann2020trajectron++} propose recurrent neural network models that predict distributions over future agent trajectories by learning from past timeseries of agent interactions.
Schmerling et al. \cite{schmerling2018multimodal} incorporate such trajectory forecasting models to construct control policies for AVs that anticipate the behavior of nearby human drivers to smoothly negotiate lane changes. Our work is complementary to such prior research -- rather than develop novel forecasting models, we instead develop novel \textit{learning} techniques to specialize models from heterogeneous agent interactions while protecting data privacy.

\textit{Federated Learning in Robotics: }
We note prior work has applied FL to robotics, but to learn vision models in private scenarios \cite{li2019fc,liu2019lifelong,liu2020federated}. Instead, we address \textit{personalized} federated learning to adapt to heterogenous human interactions, which is the key novelty of our paper. Moreover, a key novelty of our work is our new algorithm for personalization with differentiates between common parameters, such as for dynamics models, that apply to many robots and those that are robot, human, or scenario-specific.

\section{Problem Statement and Proposed Method}
\label{sec:proposed_method}

\begin{figure}[t]
    \centering
    \includegraphics[width=1.0\columnwidth]{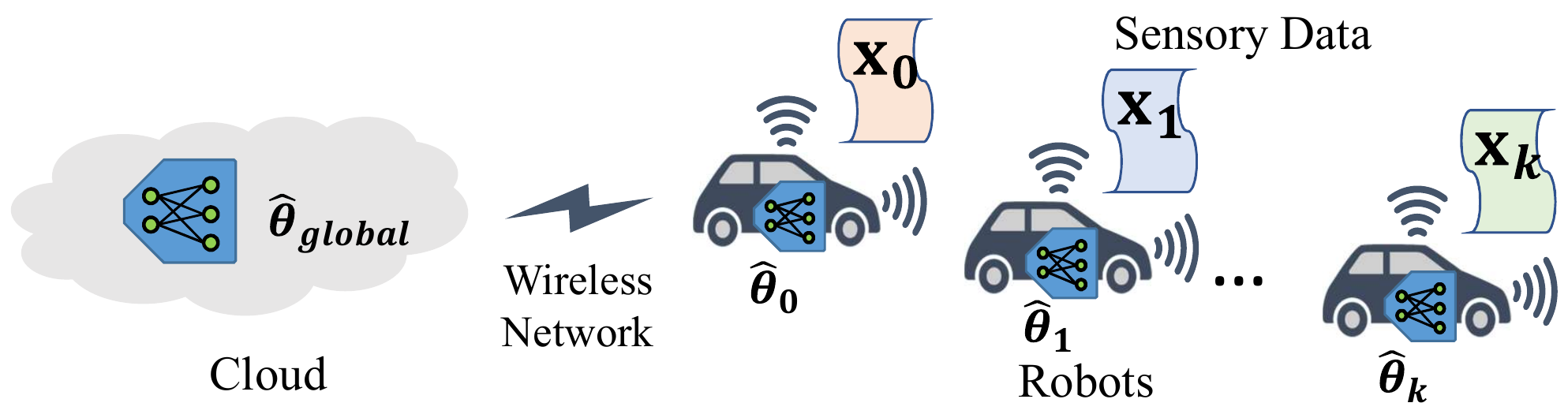}
    \caption{\small{\textbf{Personalized Federated Learning (FL) for robotics:} Each robot sees a potentially diverse local dataset $\mathbf{x}_k$ and must learn a local model with parameters $\hat{\theta}_k$. These local models are occasionally pooled and synced with a cloud server to yield a global model $\hat{\theta}_{\mathrm{global}}$ which is further specialized locally at each robot $k$.}}
    \label{fig:problem_statement}
\vspace{-1em}
\end{figure}

We now formalize our problem statement to show how FL is applied to robotic systems. 
Fig. \ref{fig:problem_statement} illustrates an overview of the system we assume in our problem.
First, we introduce a fleet of robots, such as autonomous vehicles, consisting of $n$ robots denoted by $\mathcal{R}_k (k = 0,1,..,n-1)$ which measure sensory data $x^t_k$ such as an image or LIDAR point cloud where $t \in \mathbb{N}$ is discrete time.
$\mathcal{R}_k$ connects to a cloud environment $\mathcal{C}$ with a wireless network. Our goal is to learn a model that predicts future system states or controls of other objects, such as pedestrians and vehicles, based on measured data $x^t_k$ and a parameter $\theta_k$ to be learned.
Additionally, we introduce $\theta_\mathrm{global}$ which is a parameter of the global model.
The global model is held in the cloud $\mathcal{C}$ and trained without sharing the measured data with robots.

The measured dataset is used to learn a machine learning model, such as a trajectory forecasting model,
that predicts future values $x^t_k$ with model parameters $\theta_k$.
A series of $x^t_k$ measured from time $0$ to $T$, such as a trajectory of a specific object or a segment of video, is denoted by $\mathbf{x}_k = x^{0:T}_k$.
We note that our problem especially applies when each $\mathbf{x}_k$ is very different from others.
This is because personalization is not needed if $\mathbf{x}_k$ has the same distribution for all $k$.

The learning algorithms of our federated learning system aim to learn the parameters of a prediction model.
Therefore, similar to general ML algorithms, we assume an objective function $F$ to be optimized.
Typically, this is a loss function of a prediction model, such as mean squared error for future predicted system states.
For personalization, the learning algorithms aim to find optimal parameter $\theta^*_k$ which minimizes $F(\mathbf{x}_k;\theta_k)$ for each robot $\mathcal{R}_k$. Thus, we formalize our problem as a minimization problem of the sum of the objective functions $F(\mathbf{x}_k;\theta_k)$:
\begin{small}
\begin{align*}
    \min{\sum^{n-1}_{k=0}F(\mathbf{x}_k;\theta_k)} \;\;\;\mathrm{where} \:\mathbf{x}_k = x^{0:T}_k.
\end{align*}
\end{small}
This means learning algorithms for our system should learn $\hat{\theta}_k$ that minimizes the expected value of the objective functions by training the prediction model at robot $\mathcal{R}_k$.
We note that $\hat{\theta}_k$ are the same value for all $k$ if the algorithm trains only a single global model, like general FedAVG \cite{mcmahan2017communication} without personalization. 

\subsection{Proposed Method}
Even in the setting we showed above, general FL algorithms such as FedAVG can potentially perform well.
Namely, we can obtain a better set of parameters $\hat{\theta}_k$ than randomly initialized without sharing $\mathbf{x}_k$ in a cloud environment $\mathcal{C}$.
However, we hypothesize that we can improve $\hat{\theta}_k$ by 
syncing parameters from a global model as in standard FL, but also specializing the model for robot $\mathcal{R}_k$ with the local data $\mathbf{x}_k$.
Therefore, we propose a variant of FL algorithm with an additional training step for personalization, which is shown in Alg. \ref{alg:train}.
There are two major differences from general FL.
First, we add a training step executed on each robot $\mathcal{R}_k$ for personalization in Alg. \ref{alg:train} line 7.
Unlike some prior personalized FL works such as \cite{fallah2020personalized}, we assume $\mathcal{R}_k$ can execute many more training steps (e.g. not one step of SGD but a few epochs) because today's robot systems can do so with modern low-power deep learning accelerators.
Our main innovation is to carefully limit parameter updates to avoid overfitting by personalized training.
This is inspired by standard methods in fine-tuning DNNs using transfer learning, which freeze some layers' parameters. 
Also, retraining all the parameters on a local dataset, which can be of limited size and highly biased, often causes overfitting.
We found that it is effective to slow down the update of parameters that have common features across many robots' data during personalized training.
To do so, our key insight is to estimate the variance of different parameters \textit{across} robots during FL, which is codified in Algorithm lines 7, 13, and 14.

Specifically, we now discuss how to properly adjust the extent of parameter updates.
In short, we propose to apply an adaptive learning rate to FL for robotics systems.
Generally, it is hard to accurately determine which parameters learn common dynamics and shared trends across robots,
especially when training a model such as a DNN with a huge number of parameters.
However, we can expect such common parameters tend to eventually converge to a similar value on any dataset.
Therefore, we propose a method to estimate how much parameters have in common from the variation of parameters learned by each robot in a fleet.
The estimated degree of similarity finally affects the learning rate of each parameter (lines 13-14).
As a result, higher learning rates are assigned to parameters with higher variation in personalized training, which might indicate
they need \textit{specialization} per robot.
Specifically, we introduce a variation vector $\vec{\sigma}$ whose elements represent the degree of variability of each parameter.
Then, the learning rate of each parameter is determined based on a ratio of the variability.
Namely, the learning rate of the $i$-th parameter of parameter vector $\theta$ is denoted by $L\sigma_i/\max\vec{\sigma}$, where $L$ is a constant representing the maximum learning rate.
In practical systems, we can cluster parameters (\textit{e.g.} parameters of a layer of a DNN) and calculate learning rates for each cluster for computational scalability.

\begin{small}
\begin{algorithm}[t]
\DontPrintSemicolon
\SetNoFillComment 
\LinesNumbered
\SetAlgoLined
 Randomly initialize parameter $\theta_{global}$ \;
 Initialize $\vec{\sigma} = \{\sigma_0, \sigma_1,\dots,\sigma_I\} \gets \{1,1,\dots,1\}$ \;
 Initialize $\vec{l} = \{l_0,l_1,\dots,l_I\} \gets \{L,L,\dots,L\}$ \;
 \Repeat{$\hat{\theta}_k$ converges}{
    \For{$k~ \gets0$ \KwTo $n-1$}{
        $\theta_k \gets \theta_{global} $ \;
        Update $\theta_k$ with $F(\mathbf{x}_k;)$ and learning rate $\vec{l}$\;
        $\hat{\theta}_k \gets \theta_k$, $\theta_k \gets \theta_{global} $ \;
        Update $\theta_k$ with $F(\mathbf{x}_k;)$ and learning rate $L$ \;
        Send $\theta_k$ to cloud environment $\mathcal{C}$ \;
        \KwResult{Return $\hat{\theta}_k$}
    }
    Update $\theta_{global}$ by aggregating $\{\theta_0,\theta_1,\dots,\theta_{n-1}\}$ \;
    Update $\vec{\sigma}$: $\sigma_i \gets \sum_{k=0}^{n-1}(\theta^i_k - \bar{\theta}^i)^2$ \;
    Update $\vec{l}$: $l_i \gets L\sigma_i/\max\{\sigma_0,\sigma_1,\dots,\sigma_I\}$ \;
 }
 \caption{\textbf{Federated Control Learning}:
 \small{
 $L$ is a scalar constant representing the basic learning rate,
 and $i$ is the id of an individual parameter in $\theta_k$.
 Therefore, $\theta^i_k$ represents the $i$-th parameter in $\theta_k$ (the number of individual parameters is $I$).
 The key process is line 7 where a model is trained with the adaptive learning rate $\vec{l}$.
 $\vec{\sigma}$, which is used to update $\vec{l}$, which represents a vector of the deviation sum of squares for each $\theta^i$.
 We assume every robot (client) $\mathcal{R}_k$ participates in every training round.}}
 \label{alg:train}
\end{algorithm}
\end{small}

\section{Experimental Results}
\label{sec:experiments}
\begin{figure}[t]
    \centering
    \includegraphics[width=1.0\columnwidth]{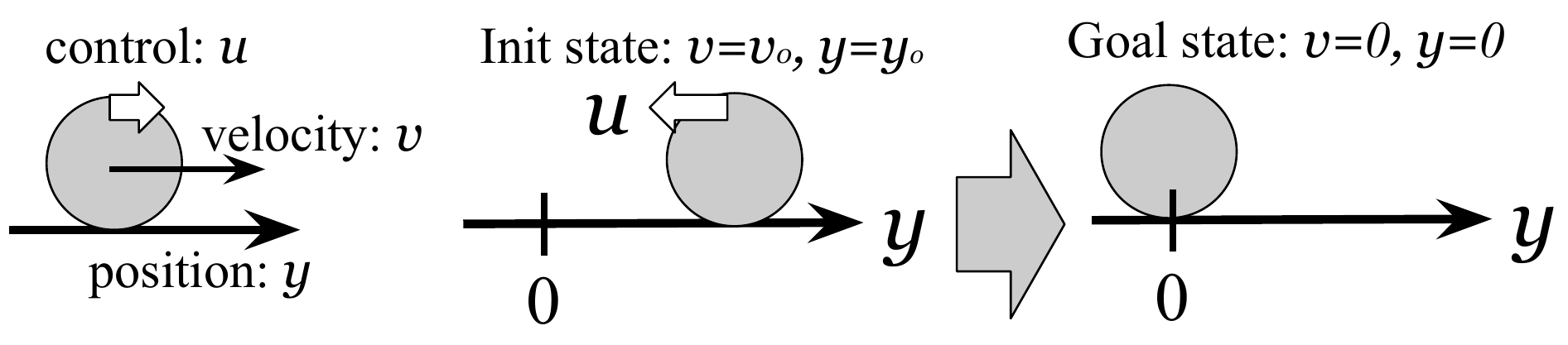}
    \caption{\small{\textbf{The motion of a mass point in 1-dimension space}:
    A mass point (circle in the fig) has a state consisting of velocity ($v$) and position ($y$), and can receive control input ($u$). (\textbf{left})
    The goal of this task is to make position and velocity zero from the init state which is randomly given. (\textbf{middle and right})}}
    \label{fig:LQR_1d_system}
\vspace{-1em}
\end{figure}

We now evaluate our proposed method (Algorithm 1) on three diverse tasks.
We start with a simple toy example to benchmark the proposed method. The second and third are more practical tasks in autonomous driving where a robot vehicle changes a lane while anticipating the behavior of other human-driven vehicles.

\subsection{Illustrative Toy Example: LQR control of a 1-dimensional point mass ``robot''}
\label{subsec:LQR_eval}

In this section, we introduce a very simple task where a point mass must be controlled by a  linear-quadratic regulator (LQR) controller in a 1-dimensional space, as illustrated in Fig. \ref{fig:LQR_1d_system}. The goal is to learn parameters of an LQR controller that moves the point mass from an arbitrary initial state to the origin. Importantly, we have $K=3$ robots with \textit{identical dynamics}, but different LQR cost functions. 
Our goal is to estimate both the dynamics model and LQR control policy parameters using local datasets $\mathbf{x}_k$ for robot $k$ and personalized federated learning to share knowledge amongst robots $k$. 
We now introduce the system dynamics and LQR controller for each robot $k$.

Each robot $k$'s system state at discrete time $t$ is denoted by $x^k_t = [y^k_t, v^k_t]$, which is a vector
of its velocity $v^k_t$ and position $y^k_t$. The dynamics of this simple linear system are given by
$x^k_{t+1} = Ax^k_t + Bu^k_t$, where $A \in \mathbb{R}^{2\times 2}$ is the dynamics matrix, $B \in \mathbb{R}^{2 \times 1}$ is the control matrix, and $u^k_t \in \mathbb{R}$ is the control input. Assuming a unit-mass system, the dynamics of our linear system are given by:
\begin{small}
\begin{align*}
x^k_{t+1} = \underbrace{\begin{pmatrix}1 & 1 \\0 & 1\end{pmatrix}}_{A} x^k_t + \underbrace{\begin{pmatrix}0\\1\end{pmatrix}}_{B} u^k_t.
\end{align*}
\end{small}
Next, we define a cost function to identify the optimal control input.
The standard LQR cost is defined as $\sum_{t=0}^{\infty} ((x^{k}_t)^\mathrm{T} Q^k x^k_t + (u^k_t)^\mathrm{T} R^k u^t_k)$, where $Q^k$ and $R^k$ are weight matrices ( $Q^k \in \mathbb{R}^{2 \times 2}, R^k \in \mathbb{R}^{1 \times 1}$ ).
It is well-known that the optimal control input $u^{*,k}_t = \mathcal{K}^{\mathrm{LQR}} x^k_t$, where $\mathcal{K}^{\mathrm{LQR}}$ is a feedback matrix that arises from the solution to the discrete time Riccati Equation.

In our toy experiment, we assume each robot $\mathcal{R}_k$ knows the parametric form of the linear dynamics equations and linear feedback policy, but does not know the specific parameters $A$, $B$, $Q^k$ and $R^k$. Instead, each robot must \textit{learn} these parameters from measured rollouts (e.g., trajectory data) of an expert controlling the system. We note that it is sufficient to learn dynamics matrices $A$ and $B$ to predict the next system state from the current state and applied control, as well as the LQR feedback matrix $\mathcal{K}^{\mathrm{LQR},k}$ to general controls from the current system state.

Crucially, since each robot has the same dynamics, our experiments should show that $A$ and $B$ quickly converge to the same global value using knowledge transfer. However, since the cost functions differ, we should see each robot converge to its unique LQR feedback matrix, illustrating the benefits of personalization.


\noindent \tu{Heterogenous Robot Datasets:}
We generated $K=3$ different synthetic datasets (one per robot) containing rollouts of the LQR system using common dynamics but different cost functions.
For simplicity, we fixed the weight matrix $Q^k$ as an identity matrix but varied $R^k$ for each robot $k$.
The first robot has the reference controller whose $R^k = 1$.
The second and third are 50 and 100, respectively, which indicate they prefer larger magnitude actuation.
During simulation, we added Gaussian noise to the system state and control inputs with a small variance of $0.01$.
Using the optimal controllers for each robot's cost function, we collected expert trajectories of a point
mass for 40 randomly initialized states and a simulation time horizon of $T=30$. Therefore, the number of total states collected was $3,600$ $(40 \times 30 \times 3)$, with $10\%$ of the data 
held-out as testing data.

\noindent \tu{Benchmark Algorithms:}
To compare to our proposed method (Algorithm 1), we evaluated five algorithms shown in Table \ref{tab:algorithms}.
These benchmarks rigorously cover a spectrum of algorithms used in practice today, ranging from purely local training without data sharing (``Local''), simply pooling all data in the cloud without privacy considerations (``Cloud''), standard FL with FedAvg (``SFL''), standard personalized FL (``SPFL''), and finally our proposed method with adaptive, parameter-wise learning rates for personalized FL (``APFL''). Of course, each robot only has access to its own training dataset locally.


\begin{table}[t]
\caption{{\textbf{Description of training algorithms}}}
\label{tab:algorithms}
\centering
\begin{tabular}{p{2.7cm}|p{5cm}}
    \hline
    Name & Description \\
    \hline \hline
    Local & All robots train their models only on data they measure themselves. \\ \hline
    Cloud & Training is processed on the cloud environment using all raw data, without privacy guarantees. \\ \hline
    Standard FL (SFL) & Standard FL, which is the same as Alg. \ref{alg:train} except there is no personalization (line 7). \\ \hline
    Standard Personalized FL (SPFL) & Standard \textit{personalized} FL, which is Alg. \ref{alg:train} without applying our contribution of a parameter-wise learning rate. \\ \hline
    Adaptive Personalized FL (APFL, \textbf{Ours}) & Our proposed method in Algorithm \ref{alg:train}. \\ \hline 
\end{tabular}
\end{table}

\noindent \tu{Evaluation Metrics:}
The overall loss function $F(\mathbf{x}_k;\theta_k)$ is the error in predicting a trajectory rollout using learned parameters $\hat{A}^k$, $\hat{B}^k$, and $\hat{\mathcal{K}}^{\mathrm{LQR,k}}$ instead of the true parameters $A$, $B$, and $\mathcal{K}^{\mathrm{LQR,k}}$. This is a sum of the mean squared error (MSE) of the control loss and state loss.
Therefore, the loss function $F(\mathbf{x}_k;\theta_k)$ is denoted by $1/|\mathbf{x}_k|\sum_{k,t}( \Vert \hat{x}^k_t-x^k_t \Vert_2^2 + (\hat{u}^k_t-u^k_t)^2)$ where $\hat{x}^k_t$ and $\hat{u}^k_t$ are prediction vectors of the system state and control input computed by $\hat{A}x^k_{t-1} + \hat{B}u^k_{t-1}$ and $-\hat{\mathcal{K}}^{\mathrm{LQR,k}} x^k_{t-1}$, respectively.
We implemented the training algorithms in Table 1 and evaluated each robot on \textit{test} data drawn from its local data distribution. Then, we report the loss averaged across each robot, which corresponds to the objective function in Sec. \ref{sec:proposed_method}. We used the ADAM optimizer \cite{kingma2014adam}, and both the default learning rate and initial uniform learning rate $L$ in Algorithm 1 are set to $0.01$.

\begin{table}[t]
\caption{\small{\textbf{Prediction loss for each algorithm}:
        We present the total loss and constituent terms of state loss and control loss for each scheme, averaged over 10 test trials. Clearly, our method of APFL has the lowest total loss (our optimization objective) since it learns the global dynamics but learns unique cost functions per robot, which is crucial to achieve low control loss. We learn the dynamics and control with a small, non-zero loss since our simulations had Gaussian dynamics and actuation noise.
}}
\label{tab:LQRloss}
\centering
\begin{tabular}{c|c|c|c|c|c}
    \hline
    Algorithm & Local & Cloud & SFL & SPFL & APFL \\ \hline
    \hline
    State Loss & 0.01101 & \textbf{0.01012} & 0.01023 & 0.01023 & 0.01018 \\ \hline
    Control Loss & 0.01251 & 0.19666 & 0.20123 & \textbf{0.01095} & 0.01096 \\ \hline
    Total Loss & 0.02352 & 0.20678 & 0.21146 & 0.02118 & \textbf{0.02114} \\ \hline
\end{tabular}
\end{table}

\begin{figure}[t]
    \centering
    \includegraphics[width=0.8\columnwidth]{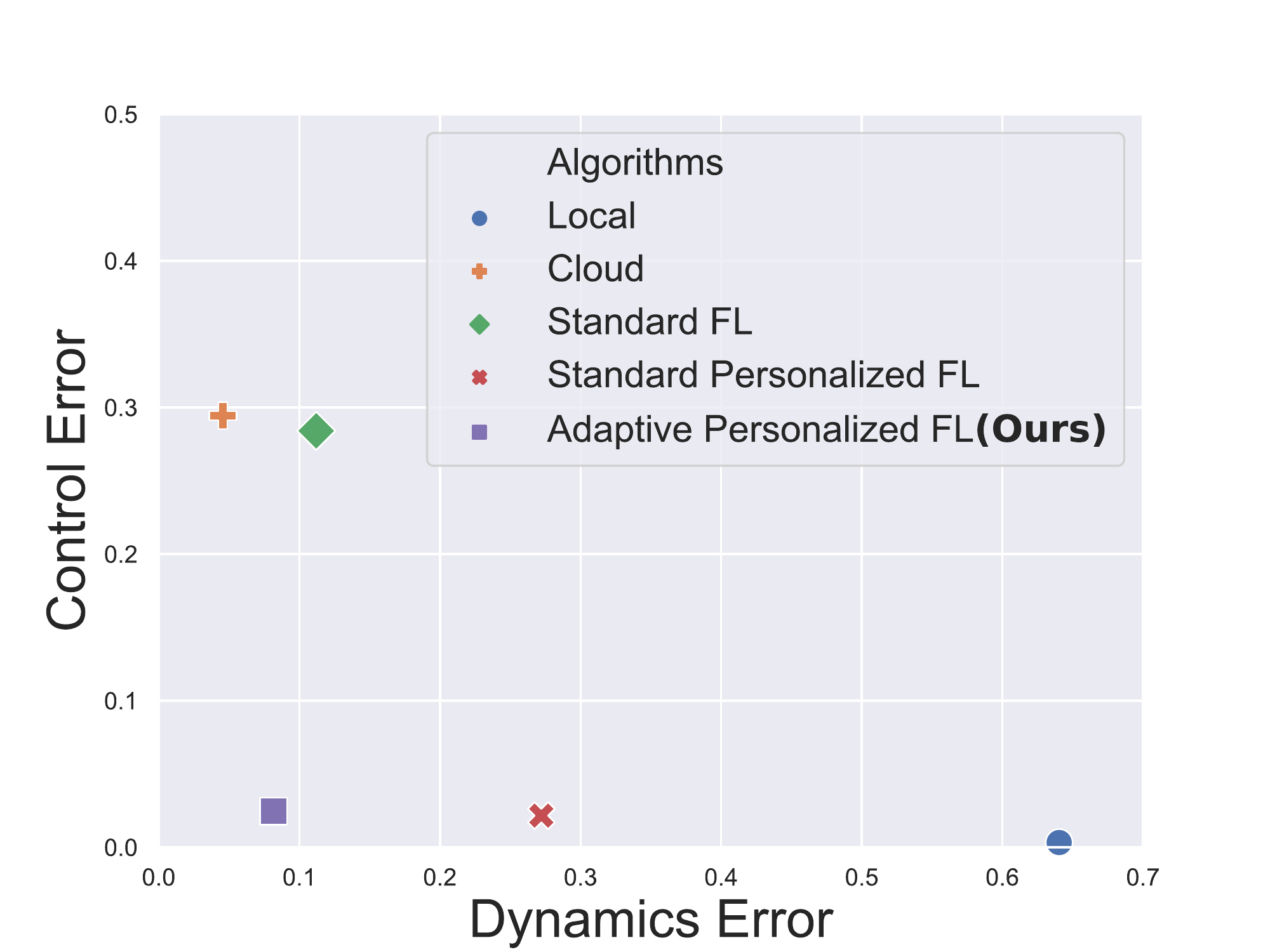}
    \caption{\small{\textbf{Divergence of learned parameters from true parameters for Toy LQR Example}:
    We quantify the divergence as the Euclidean distance between learned and true parameters
    for LQR in terms of $A$, $B$ for dynamics (x-axis) and the feedback matrix $\mathcal{K}^{\mathrm{LQR,k}}$ (y-axis).
    Clearly, our method of APFL (purple) efficiently learns global insights for the shared dynamics to achieve low error for dynamics parameters. Moreover, it also effectively specializes the model to learn local, heterogeneous cost functions and achieve low error for control parameters.
    }}
    \label{fig:LQR_distance}
\vspace{-1em}
\end{figure}

\noindent \tu{Results:}
Table \ref{tab:LQRloss} shows the prediction losses for each training method.
We show the prediction loss for next system states ($x^k_t$ predicted by $A$ and $B$) and control loss ($u^k_t$ predicted by $\mathcal{K}^{LQR,k}$) separately to investigate how learning affects shared and robot-specific parameters. Most methods are able to learn the common state dynamics that depend on the global $A$ and $B$.
However, purely local training achieves the worst loss since each robot has a smaller amount of local training data.
Our key result is that for control loss, the methods with personalization (Local, SPFL, and our APFL) outperform others that train a global model by mixing robots' heterogenous data.
Crucially, our method of APFL has the lowest total loss.

Fig. \ref{fig:LQR_distance} provides further evidence for the benefits of our APFL method
by showing how close the learned parameters $\hat{A}^k$, $\hat{B}^k$, and $\hat{\mathcal{K}}^{\mathrm{LQR},k}$
differ from the ground-truth parameters.
Clearly, our method of APFL (purple) shows strong benefits of not only personalization, but also our scheme of adaptively adjusting the learning rate of parameters based on their variation across robots. Specifically, we achieve both low errors for dynamics and control matrices (x-axis) and the estimated LQR feedback matrix (y-axis). 
Next we evaluate the benefits of APFL on two challenging AV
tasks featuring rich human-robot interaction in our user studies.


\subsection{Lane swapping in the CARLO driving simulator}
\label{subsec:lane_swapping}

\begin{figure}[t]
    \centering
    \includegraphics[width=1.0\columnwidth]{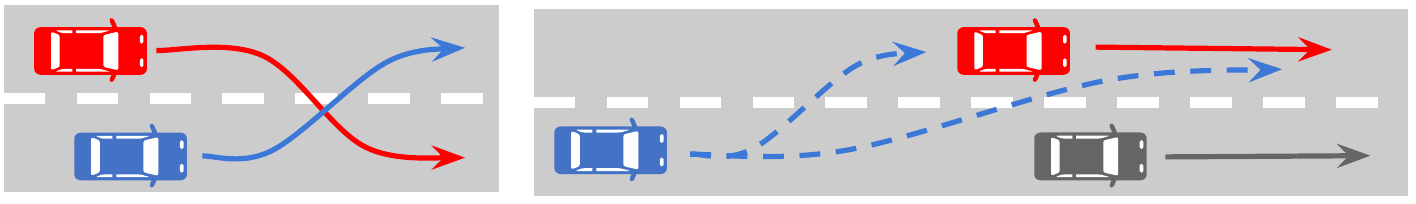}
    \caption{\small{\textbf{Traffic scenarios}:
    The left and right figures show scenarios tested in Sec. \ref{subsec:lane_swapping} and Sec. \ref{subsec:lane_change}, respectively. 
    The red car (upper lane) is a human-driven car, and the blue car (bottom lane) is an AV.
    The AV uses a forecasting model, learned using our personalized FL method, to 
    anticipate the future trajectory of the human-driven car (for heterogenous driver styles) to efficiently and safely change lanes.
    In the right scenario, the gray car (lower right) is always going straight but at variable distances from the blue AV, 
    requiring the AV to deliberate where it has a window to swap lanes.
    }}
    \label{fig:traffic_scenarios}
\vspace{-1em}
\end{figure}

The second case study is shown on the left side of Fig. \ref{fig:traffic_scenarios},
where a human-driven vehicle and AV must safely interchange lanes in a short distance, inspired by \cite{schmerling2018multimodal}.
The AV is equipped with a trajectory forecasting model that predicts the future motion of the human-driven car, conditioned on past interaction history and a candidate robot future control decision. 
A key feature of our work is we performed a study with 7 real human drivers, who exhibited a mixture of aggressive and cautious driving styles when deciding whether to overtake or yield to the AV. To quickly collect data with diverse human volunteers, we used the CARLO 2D driving simulator \cite{cao2020reinforcement}, which is a light-weight version of CARLA without photo-realistic rendered scenes.
Crucially, our FL framework ensures that the human-robot interaction datasets, which often show risky driving styles from human subjects, are kept private and never shared with a central server.


\noindent \tu{System Definition:}
We define the joint robot and human (i.e., system) state $S_t$ as $(\vec{p}^{\;r}_t, \vec{p}^{\;h}_t, \vec{v}^{\;r}_t, \vec{v}^{\;h}_t)$ where $\vec{p}^{\;r}_t$ and $\vec{p}^{\;h}_t$ are the 2D position vectors of the robot and human car at discrete time $t$, and $\vec{v}^{\;r}_t$ and $\vec{v}^{\;r}_t$ are the velocity vectors.
The 2D vectors are composed of $x$ and $y$ components.
For example, a robot's velocity is denoted by $\vec{v}^{\;r}_t = (v^{rx}_t, v^{ry}_t)$ where $v^{rx}_t$ and $v^{ry}_t$ are scalar values representing a robot's velocity in the lateral and longitudinal directions.
The control space of each car involves throttle and steering
which are finite sets. 
For example, the throttle of a robot is denoted by $c^{rth}_t \in \{-1.5,0,1.5\}$.
Similarly, the steering of a human is denoted by $c^{hst}_t \in \{-0.04,0,0.04\}$.
Thus, the control of the cars at time $t$ is denoted by $C_t = (C^r_t, C^h_t) = (c^{rth}_t, c^{rst}_t, c^{hth}_t, c^{hst}_t)$.

\begin{figure}[t]
    \centering
    \includegraphics[width=1.0\columnwidth]{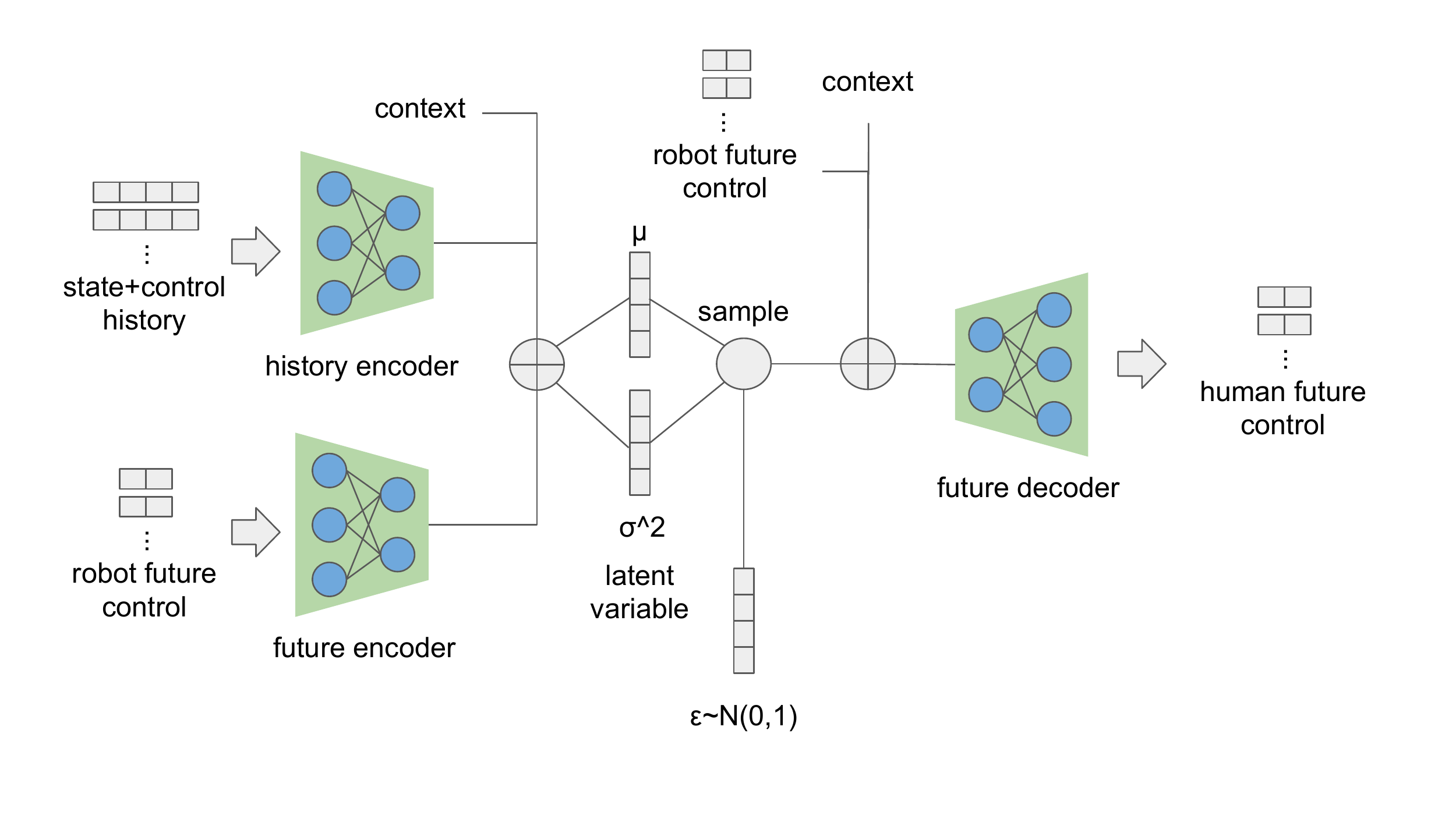}
    \caption{\small{\textbf{Architecture of prediction model}:
    We learn a variational autoencoder for trajectory forecasting using LSTM cells for the encoder and decoder
    \cite{schmerling2018multimodal} using our personalized FL method.
    The inputs are 1) a history of past system states and controls, 2) a candidate robot future control and 3) an \textit{optional} indicator of context (such as location).
    The encoder maps 1) and 2) to latent variables $\mu$ and $\sigma^2$.
    The decoder generates a prediction of the human-driver's trajectory from the latent variables, 2) and 3).
    }}
    \label{fig:cvae_arch}
\vspace{-1em}
\end{figure}

\noindent \tu{Prediction Models for Federated Learning:}
The robot controller has to make predictions of future system states to choose appropriate controls.
Our scenario is inspired by \cite{schmerling2018multimodal}, and makes the
same practical assumption that we know the AV's internal dynamics and can fully observe
robot and human states. Crucially, all we need is to predict the human's future control inputs
and trajectory \textit{conditioned} on candidate future robot controls. Then, given such a prediction model,
we can embed it into a Model Predictive Controller (MPC) \cite{MPC} for the AV to choose low cost robot actions
that \textit{anticipate} the response of humans. This means that the prediction model needs to estimate a future \textit{series} of human control inputs at a single time-step. To achieve this, we introduce a Conditional Variational Autoencoder (CVAE) illustrated in Fig. \ref{fig:cvae_arch} with recurrent subcomponents \cite{schmerling2018multimodal} which are able to predict time-series data.
Our MPC control inputs are decided periodically at a discrete interval $\tau$.


\noindent \tu{Robot Controller:}
We define the robot controller as a two-phase controller.
In the first phase, the controller is responsible for negotiating the longitudinal distance between human vehicles with different driving styles.
This is the most important part in our experiment because it requires accurate, personalized predictions of human driver behavior.
For the first phase, we introduce a cost function $J_t$ to evaluate candidate control inputs at time $t$ as follows:
\begin{small}
\begin{align*}
    J_t = \alpha (p^{ry}_{t} - p^{hy}_{t}) (v^{ry}_{t} - v^{hy}_{t}) + \beta / |\vec{p}^{\;r}_{t} - \vec{p}^{\;h}_{t}|
\end{align*}
\end{small}
The first term represents longitudinal cost based on the fact that cars need sufficient longitudinal distance during a lane change.
The second term represents a distance cost between cars to avoid collisions.
$\alpha$ and $\beta$ are parameters that weight each objective. We chose $\alpha=-3$ and $\beta=5000$ 
to heavily emphasize safety in this experiment.

When the robot controller decides a series of control inputs $[(c^{rth}_{t+1}, c^{rst}_{t+1}), (c^{rth}_{t+2}, c^{rst}_{t+2}),..,(c^{rth}_{t+\tau}, c^{rst}_{t+\tau})]$ at time $t$, it computes the sum of the expected cost function in the time horizon $H > \tau$ for all possible robot control candidates.
The controller chooses a series of control inputs that minimizes the sum of the expected cost, which naturally depends on anticipated future human controls.
We set the control interval $\tau$ and planning horizon $H$ to $3$ and $20$ respectively, which performed well with an acceptable compute budget.

Once there is enough longitudinal distance to swap lanes, the controller switches to the second phase, and just works as a lane change controller.
Since we observed humans' driving styles almost do not differ during the lane change (and largely govern \textit{when} to safely initiate it),
the lane change controller is a static LQR controller. This static LQR controller drives the robot to  
a target lateral position $p^{rx}$ representing the x-coordinate of a target lane with longitudinal velocity $v^{ry}$.
In the following quantitative evaluation, we incorporate human trajectory prediction models trained with the benchmarks in Table 1
into the MPC-based AV controller. Then, we compare these controllers in terms of how safe and efficiently the cars are able to swap lanes given the different human's driving styles.

\noindent \tu{Dataset:}
Our dataset consists of system trajectories (including robot and human states, controls, and costs) from time $0$ to the time when both cars finished swapping lanes.
The dataset consists of a training set, a held-out evaluation set, and a third held-out test set.
In order to add diversity to each session's data, we randomly created 50 initial states $S_0$.

For the training set, seven humans play the lane swapping scenario running on the CARLO simulator with 50 initial states above.
Since the prediction model with CVAE shown in Fig. \ref{fig:cvae_arch} is not trained at this time,
the robot controller uses a very simple model based on only the current state.
Specifically, the predicted human control $\hat{C}^h_i$ for future time $i$ is always the current human control $C^h_t$ for any $i \in \{t+1,t+2,..,t+\tau\}$.

Once we gather initial data for how humans respond to a baseline AV without a sophisticated model of human behavior, we can train a better driven prediction model. This prediction model
is in turn used in the AV's MPC and resulting rollouts against human drivers allows us to generate an evaluation set. Notably,
this evaluation set is used to fine-tune and re-train the human trajectory forecasting model.

Similar to the experiment \ref{subsec:LQR_eval}, each of the 7 robot cars observes only one type of data corresponding to one human driver.
During training, both the default learning rate and $L$ are 0.001, and the loss function is the evidence lower bound objective (ELBO) \cite{kingma2013auto,rezende2014stochastic}.
The number of epochs is 30 for each training step in Alg. \ref{alg:train}.
We used 40 sessions for training and 10 challenging held-out sessions for testing.
We trained the model with four algorithms shown in Tab. \ref{tab:algorithms}.
Similar to the experiment in Sec. \ref{subsec:LQR_eval}, the federated learning algorithms use FedAVG to aggregate trained parameters.
In our method (APFL), we used layer-wise learning rate instead of parameter-wise to improve compute efficiency, which essentially
groups parameters in one layer.

\begin{figure}[t]
    \centering
    \includegraphics[width=1.0\columnwidth]{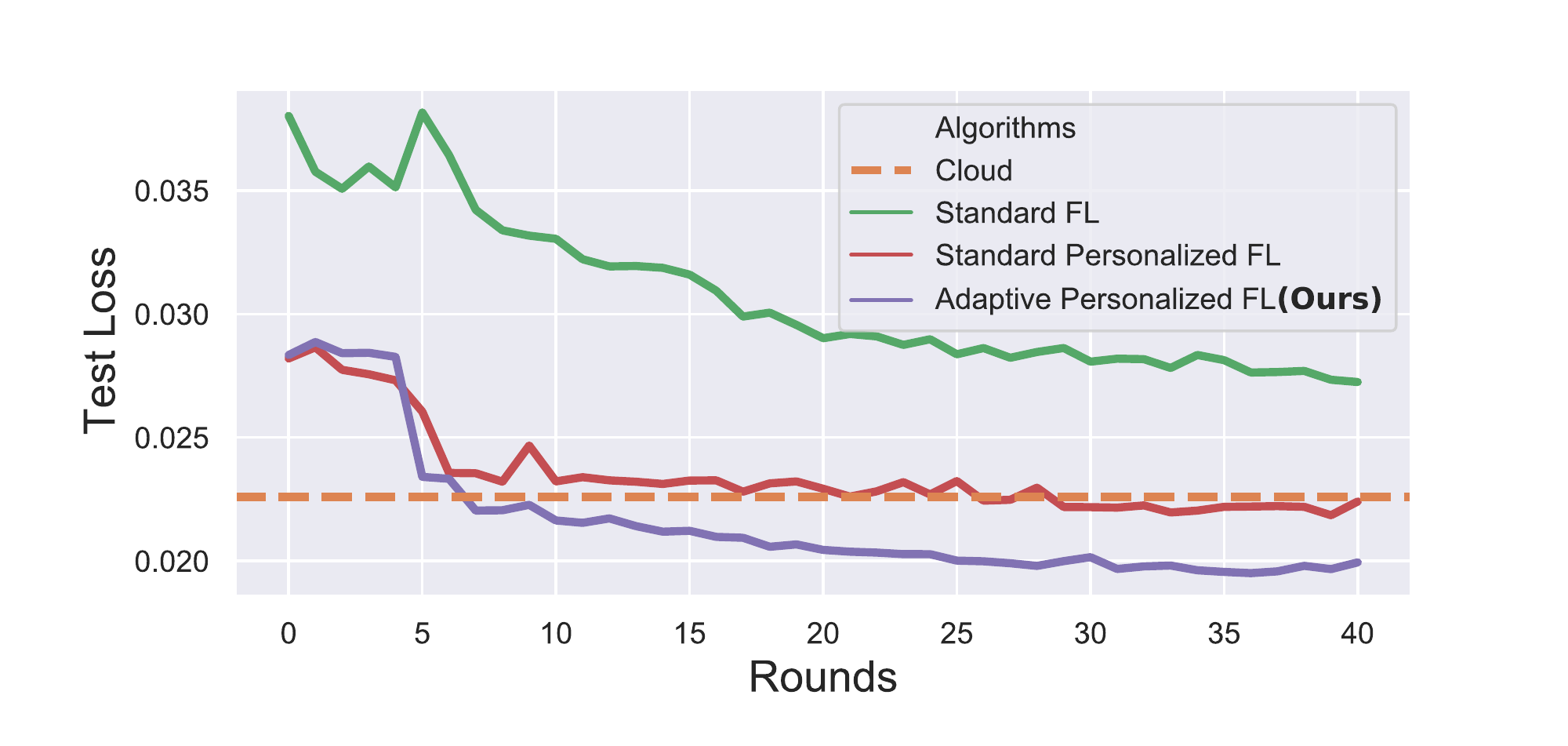}
    \caption{\small{\textbf{Learning curve of FL algorithms in the lane swapping scenario}:
    Clearly, our method of APFL (purple) achieves a lower test loss that is personalized to heterogenous drivers in our user study for a lane merging scenario in CARLO.}}
    \label{fig:carlo_lc}
\vspace{-1em}
\end{figure}

\noindent \tu{Evaluation Results:}
We evaluate the test losses of the prediction models trained with our benchmark algorithms.
Fig. \ref{fig:carlo_lc} illustrates the held-out \textit{test} dataset loss for model's successively trained using
our benchmark algorithms. In contrast to standard FL, algorithms involving personalization adapt better
to heterogenous human driver interactions to achieve a lower loss. Crucially, our method of APFL achieves the lowest loss, which is consistent with the results of the experiment in Sec. \ref{subsec:LQR_eval}.

\begin{figure}[t]
    \centering
    \includegraphics[width=1.0\columnwidth]{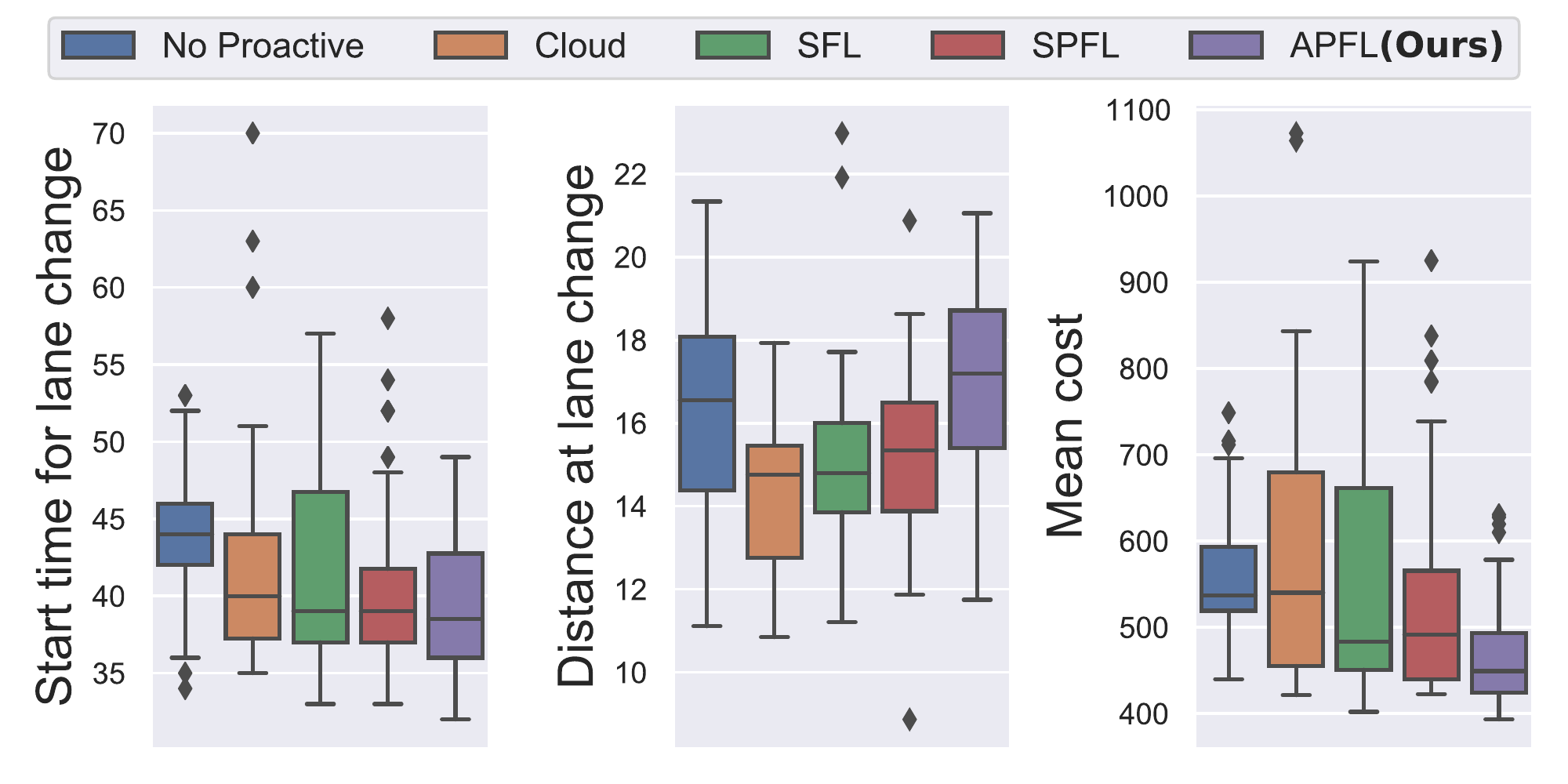}
    \caption{\small{\textbf{Performance comparison of lane swapping scenario}:
    The non-proactive scheme (blue) means a simple benchmark case where the robot controller uses a na\"ive constant-velocity prediction model. 
    APFL shows statistically-significant benefits from the other training algorithms in mean cost. Indeed, the maximum p-value for the Wilcoxon Signed-Rank test was 0.002476 when we compare the closest competitor of SPFL with APFL. Moreover, our method of APFL maintains a larger safe distance between cars.
    }}
    \label{fig:carlo_boxplot}
\vspace{-1em}
\end{figure}

Additionally, evaluation results show an AV (with our learned prediction models in-the-loop) can safely and efficiently swap lanes with actual human drivers in \textit{test} episodes in the CARLO simulator.
Fig. \ref{fig:carlo_boxplot} shows distributions of key quantitative metrics over the test episodes, such as the elapsed time until the AV starts a lane change, distance between cars when lane changes commence, and the mean cost for each session.
Clearly, an AV using our APFL model can start to change lanes quickly with a larger safe distance to the human car.

\subsection{Lane change with the CARLA AV simulator}
\label{subsec:lane_change}

We now consider a more sophisticated AV driving scenario in the photo-realistic, standard
CARLA \cite{Dosovitskiy17} simulator. As shown on the right of Fig. \ref{fig:traffic_scenarios}, 
our scenario is similar to the lane-swap scenario of the previous section. However, there is additional complexity
since there is a gray human-driven car that starts at random distances from the AV with random velocities, which
requires the AV to reason about whether there is enough safety margin to overtake the red car.


\noindent \tu{Experimental Setup:}
The settings of this experiment closely follow those of the lane swapping case in \ref{subsec:lane_swapping}.
The control space is the same as before, and the state space is the same except that we add the state of the gray car.
The architecture of the prediction model is exactly the same as the one in \ref{subsec:lane_swapping}, shown in Fig. \ref{fig:cvae_arch}.
Further, the robot controller is also the same except it does not switch to the second phase where the robot is going to change a lane because robot cars do not change lanes in the scenario.
Additionally, we set a lower limit $C_{low}$ on the cost function to prevent the robot cars from unnecessarily increasing the distance to the human car after both cars have a safe and sufficient distance, which will artificially decrease the cost.
Therefore, the new cost function $J^\prime_t$ is denoted by $\max (C_{low}, J_t)$.

Unlike the lane swapping case, we use a programmed synthetic controller for human cars to test on uniformly diverse driving styles, including risky behaviors.
This synthetic human controller is also a two-phase controller, but control inputs are decided according to a target velocity in the first phase.
The target velocity is chosen as high $v_\mathrm{high}$ or low $v_\mathrm{low}$, where the high velocity $v_{\mathrm{high}}$ is only chosen if the human car has a safe relative difference from the other cars, modulated by its risk tolerance parameter $\gamma$: 
\begin{small}
\begin{align*}
    v^{\mathrm{target}}_t = \left\{
\begin{array}{ll}
    v_{\mathrm{high}} & \text{if}\;\; \gamma D_{\mathrm{rel}} \geq 0 \;\land\; p^{hy}_{t} - p^{gy}_{t} \geq D_{\mathrm{safe}} \\
    v_{\mathrm{low}} & \text{otherwise}
\end{array}
\right.
\end{align*}
\end{small}
In the above equation, $\gamma$ is a parameter representing the degree of preference for overtaking robot cars, which we henceforth
called the risk-tolerance parameter.
Further, in the above equation, $D_{\mathrm{rel}}$ indicates the relative gap between cars based on the positions and velocities denoted by $(p^{hy}_{t} - p^{ry}_{t}) + (v^{hy}_{t} - v^{ry}_{t})$,
and $p^{gy}_t$ and $D_{\mathrm{safe}}$ are a longitudinal position of the gray car and a safety distance constant, respectively.

The target velocity for the human-driven vehicle is set depending on the relative velocity to the robot car driving alongside it.
Once the target velocity is determined, the synthetic human-driver controller computes control inputs with a PID controller in CARLA.
We tested 5 kinds of driving styles whose risk-sensitivity ($\gamma$) parameters were in the range $[-1.0,-0.5,0,0.5,1.0]$.
These driving styles include a mixture of cautious and aggressive styles to ensure a diverse dataset.
For instance, a human car with $\gamma = 1.0$ sometimes tries to pass the robot car even when the current relative position to the robot car is minimal.

We collected a diverse dataset in the same manner as in the previous scenario.
Specifically, we generated 54 initial states where the relative position between cars and velocities are uniformly distributed.
We trained trajectory forecasting models according to Table \ref{tab:algorithms} with the same hyper-parameters and LSTM encoder-decoder model as the previous scenario. Finally, we integrated the trained trajectory forecasting models into the AV's controller and evaluated it in 9 challenging scenarios
where the robot cars need to change their behavior depending on humans' driving styles.

\begin{table}[t]
\caption{\small{\textbf{Prediction loss for each training scheme}:
All losses are on the test data after training has converged.
}}
\label{tab:lane_change_loss}
\centering
\begin{tabular}{c|c|c|c|c}
    \hline
    Algorithm & Cloud & SFL & SPFL & APFL\textbf{(Ours)} \\ \hline
    \hline
    Loss & 0.01643 & 0.01760 & 0.01600 & \textbf{0.01160} \\ \hline
\end{tabular}
\end{table}

\begin{figure}[t]
    \centering
    \includegraphics[width=1.0\columnwidth]{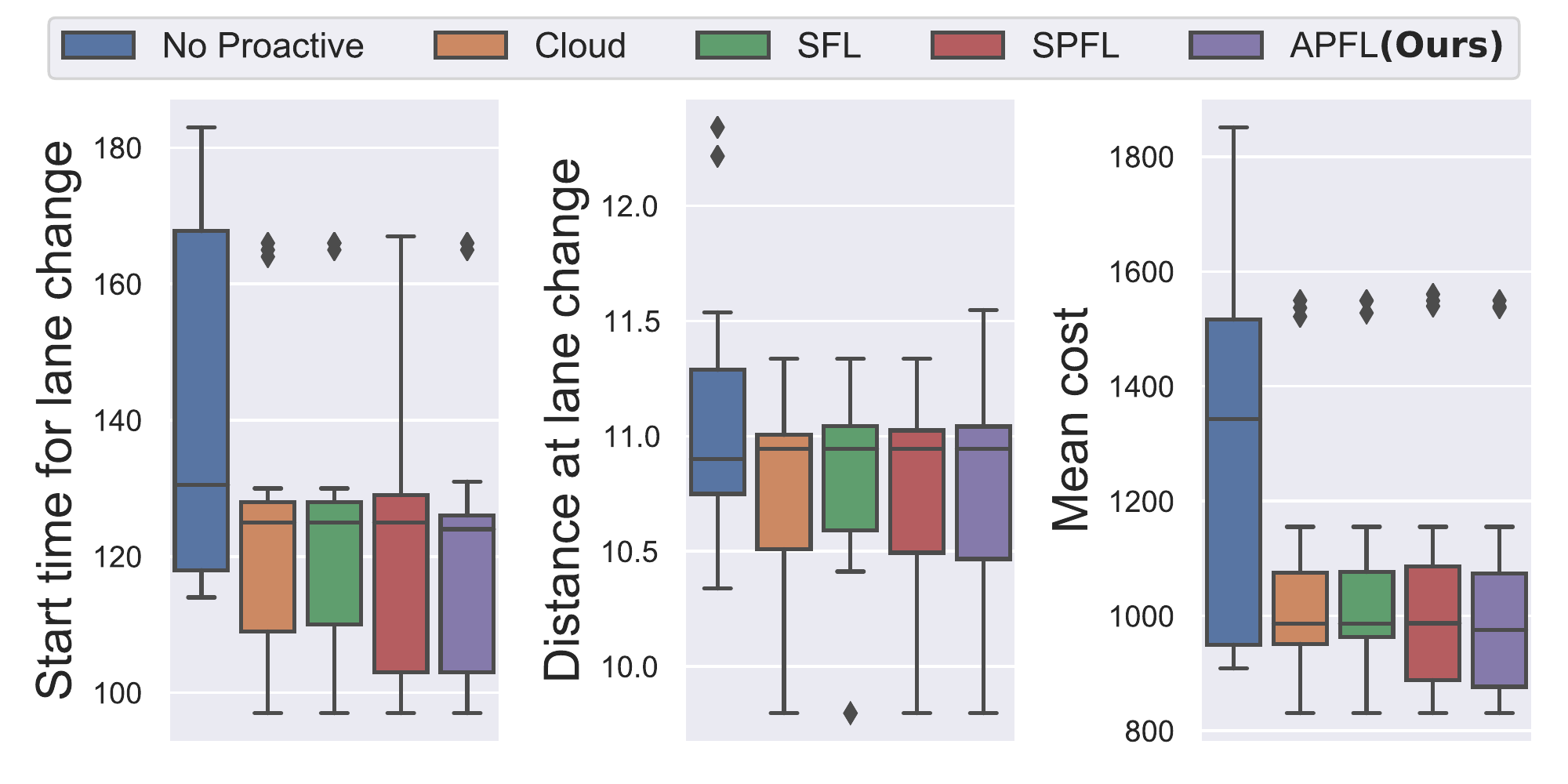}
    \caption{\small{\textbf{Performance comparison of CARLA lane change scenario}:
    Similar to the result of the lane swapping case in Sec. \ref{subsec:lane_swapping}, APFL shows significant differences from the other training algorithms in mean cost (the max p-value for the wilcoxon signed-rank test was 0.021346 for the comparison with SPFL) while maintaining a safe distance between cars when lane changes commence.
    }}
    \label{fig:carla_boxplot}
\vspace{-1em}
\end{figure}

\noindent \tu{Evaluation Results:}
Using the same metrics as the lane swapping case in Sec. \ref{subsec:lane_swapping}, we evaluate the test loss of the prediction model and the controllers' performance in terms of how safe and efficiently the cars are able to change a lane.
The prediction losses on test data for each model, which are shown in Tab. \ref{tab:lane_change_loss}, have the same trend as in the previous scenario of lane swapping, clearly illustrating the benefits of our APFL method.
Moreover, Fig. \ref{fig:carla_boxplot} illustrates APFL provides the lowest cost while maintaining a safe distance from other cars.

\section{Discussion and Conclusions}
\label{sec:conclusion}
 This paper presents a novel personalized federated learning framework for 
deployments of robots that measure diverse sensory streams in varied environmental contexts.
Our first contribution is to show the drawbacks, and potential, of standard personalized FL through a case study where robotic models
share a common structure for dynamics but heterogeneous cost functions. Then, our second contribution is to propose a novel algorithm which mitigates these drawbacks and effectively leverages both local and global knowledge to improve robotic control. Specifically, we demonstrate strong experimental performance of our algorithm in state-of-the-art driving simulators featuring real human driver data.
 
 In future work, we plan to test our algorithm on large-scale deployments of AVs using a combination of public trajectory datasets from Waymo, Aptiv, Lyft, etc. Moreover, we plan to provide theoretical convergence guarantees for our personalized FL algorithm in distributed convex optimization settings that arise in multi-agent systems and robotics. Overall, our work is a timely first step to address how to learn privacy-preserving, specialized deep learning models for robot fleets that will
 increasingly interact with diverse humans in diverse operating scenarios.
 

%

\bibliographystyle{abbrv}
\bibliography{ref/ms}

\end{document}